# Uncovering Political Hate Speech During Indian Election Campaign: A New Low-Resource Dataset and Baselines

Farhan Ahmad Jafri[1], Mohammad Aman Siddiqui[1], Surendrabikram Thapa[2],
Kritesh Rauniyar[3], Usman Naseem[4], Imran Razzak[5]

[1] Jamia Millia Islamia, India
[2] Department of Computer Science, Virginia Tech, USA
[3] Department of CSE, Delhi Technological University, India
[4] School of Computer Science, The University of Sydney, Australia
[5] School of Computer Science and Engineering, University of New South Wales, Australia

**Abstract**

The detection of hate speech in political discourse is a critical issue, and this becomes even more challenging in low-resource languages. To address this issue, we introduce a new dataset named IEHate, which contains 11,457 manually annotated Hindi tweets related to the Indian Assembly Election Campaign from November 1, 2021, to March 9, 2022. We performed a detailed analysis of the dataset, focusing on the prevalence of hate speech in political communication and the different forms of hateful language used. Additionally, we benchmark the dataset using a range of machine learning, deep learning, and transformer-based algorithms. Our experiments reveal that the performance of these models can be further improved, highlighting the need for more advanced techniques for hate speech detection in low-resource languages. In particular, the relatively higher score of human evaluation over algorithms emphasizes the importance of utilizing both human and automated approaches for effective hate speech moderation. Our IEHate dataset can serve as a valuable resource for researchers and practitioners working on developing and evaluating hate speech detection techniques in low-resource languages. Overall, our work underscores the importance of addressing the challenges of identifying and mitigating hate speech in political discourse, particularly in the context of low-resource languages. The dataset and resources for this work are made available at https://github.com/Farhan-jafri/Indian-Election.

## Introduction

In today's world, social media has become increasingly popular for expressing opinions and engaging in online discussions. Social media has turned out to be a key influence in amplifying economic, political, and cultural disparities globally, thereby impacting politics in democratic as well as autocratic regimes (Zhuravskaya, Petrova, and Enikolopov 2020). Opinions on social media vary on a vast spectrum of emotions. During events such as elections, the users on the platforms tend to be more active by voicing support or criticism for the people and parties contesting in the election. The ease of access and vast reach of social media combined with some degree of anonymity options has facilitated a hostile and offensive environment on it (Beddiar, Jahan, and Oussalah 2021). During elections, supporters of respective parties tend to gain support on social platforms by criticizing their opposition through accusations, highlighting issues, and failures. Politics is impossible to separate from sensitive topics such as Religion, Sex, Race, Culture, etc.; hence political competition becomes a forefront for hateful content online. Hate speech encompasses various types of language, such as name-calling, defamation of political rivals, false accusations, targeted sarcasm, references to historical wrongdoings, and more. Minimizing hate speech is thus important because it harms individuals and groups, fosters division and prejudice, and erodes social cohesion (Parihar, Thapa, and Mishra 2021).

India is a country of 1.21 billion people (Chandramouli and General 2011). With such a vast population and enormous diversity, hate speech on online platforms is an inevitable issue that needs to be tackled. In the Indian General elections of 2009, social media was first used to engage voters politically. In the General Elections of 2014 and 2019, the strategic use of social media by the BJP prompted other political parties to adopt it as part of their political campaigns (Mir and Rao 2022). This fierce contest on social media among the parties led to accusations and defamation among political leaders and ordinary citizens. The subtle religious difference in India have been exploited in the past and have resulted in communal violence. In the digital age, the propagation of hateful religious views has also been accelerated by the circulation of fake news and hate speech on digital platforms, serving as propaganda to influence voters' decision-making (Parwez 2022).

According to the 2011 Census, there are 43.63 percent of Indians that speak Hindi as their mother tongue (Chandramouli and General 2011). Hindi is based on the Devanagari script, which comprises 47 characters with 33 consonants and 14 vowels (Templin 2013). With the vast usage of the Hindi language on social media, the usage of NLP techniques for a low-resource language like Hindi can aid to minimize the problem of Hate Speech. To create a bridge to combat hate speech and invite more research in hate speech detection in the Hindi language, we have created "IEHate", a dataset of 11,457 Tweets collected during state assembly elections in the states of Goa, Uttarakhand, Uttar Pradesh, Punjab, and Manipur, all conducted in the year 2022. The tweets are manually annotated as "Hate" and "Non-Hate" by four annotators. Our contributions are as follows:

- We create and release a new low-resource dataset of 11,457 Hindi tweets labeled for the presence of hate speech during Indian assembly election campaign.
- Our experimental analysis shows that state-of-the-art transformer-based models are important and further research can be done in this direction.
- Our analysis shows that the algorithms showed lower performance compared to human evaluation for detecting instances of hate speech. These findings emphasize the need for developing more advanced and robust automated systems for hate speech detection during the Indian election campaign.

## Related Works

Several studies have been conducted to detect hate speech on social media platforms. Vidgen and Yasseri (2020) annotated 109,488 tweets for the identification of Islamophobic hate speech, categorizing the annotation into none, weak, and strong. Similarly, Waseem and Hovy (2016) collected 16,914 tweets which had three labels as racist, sexiest, and none. They also showed the gender distribution for the detection of hate speech. Meanwhile, Vadesara and Tanna (2023) annotated 10,000 tweets for hate speech collected using keywords such as religion, sport, politics, and celebrity, categorizing them as hate or non-hate. Mulki et al. (2019) compiled 5,846 Arabic-language tweets about political events and categorized them as regular, abusive, or hateful. These studies demonstrate the widespread use of hate speech in relation to political issues across many languages and the importance of developing effective detection methods.

In addition to the above works, several studies have been conducted on hate speech detection in the Hindi and Hindi code-mixed languages. Bohra et al. (2018) developed a dataset of 4,575 Hindi-English code-mixed tweets manually annotated for hate speech and normal speech. Similarly, Kumar et al. (2018) proposed a multilingual dataset containing Hindi, English, and mixed tweets annotated for aggression. Similarly, Modha et al. (2020) curated the data from Twitter and Facebook where they extracted code-mixed Hindi language as well as English. The task was set for online aggression detection where they used 15,001 data samples to train the models. Sreelakshmi, Premjith, and Soman (2020) made use of three existing data sources to create a pool of 10,000 data samples with two classes (Hate and Non-hate). The performance of various models was compared using different word embeddings like word2vec, fastText, and doc2vec.

Despite these efforts for the detection of hate and aggression in Hindi and Hindi-English code-mixed data, there is a need for more research and data specifically focused on detecting hate speech in political discourse in the Hindi language. As political discourse is a crucial aspect of any democracy, the spread of hate speech in this domain can have serious consequences on society. Realizing this need, our dataset, IEHate can be used to train machine learning models for hate speech detection in political discourse. By developing such models, it will be possible to monitor and control the spread of hate speech in political discourse, thus promoting a healthier and more inclusive democracy.

## Dataset

On March 10, 2022, the results of the state election in India were announced after the votes had been counted. To investigate the role of hate speech in political discourse leading up to the election, we collected tweets using the Twitter API between November 1, 2021, and March 9, 2022, which covered the period of the election campaign. We used Twitter API[1] to gather the tweets from this time period.

### Data Collection

In order to find relevant tweets, we used various hashtags like #UttarPradeshElections2022, #GoaElections2022, #PunjabElections2022, #IndianElections2022, #StateElections2022, #AssemblyElections2022, #UttarakhandElections2022 and #ManipurElections2022.

We used tweets that predominantly had Hindi text. In order to filter the tweets for our dataset, we set criteria that included retaining tweets that contained a few non-Hindi words or phrases as long as the majority of the tweet was in Hindi. However, we eliminated tweets that were deemed non-informative, such as spam or advertisements, and those that used election-related hashtags only for spamming. Additionally, we excluded tweets that lacked clear context related to the assembly election and might be influenced by local contexts, as they could introduce ambiguity and hinder the accurate categorization of hate speech. After applying these criteria, we were left with 11,457 labeled tweets that were annotated for their text.

### Annotation

The dataset was manually annotated by a team of four individuals with diverse educational backgrounds and experience in NLP and data collection. The annotators were from different regions in India and had different political and religious beliefs. Having such diverse backgrounds helps minimize bias, which is crucial in ensuring accurate labeling (Vargas et al. 2022).

The data was annotated into two binary classes, namely hate speech and no hate speech, to categorize the tweets. To accomplish this, annotators were provided with guidelines to assign labels to the tweets. In cases where the annotators were unsure about the label for a tweet, it was marked as 'Non-Informative' and eventually excluded from the dataset. The annotation guidelines are mentioned below.

---

[1] https://developer.twitter.com/en/docs/twitter-api

**Hate Speech**: Hate speech during political events such as election campaigns often targets specific groups based on their political beliefs or affiliations and expresses hostility or aggression towards them. It can also include the use of satires to disseminate harmful messages that are intended to demean, degrade or dehumanize a particular political group or individual.

> ➢AAP की सोच पाकिस्तान जैसी. पंजाब के ख़िलाफ़ साज़िश रची गई. वोट से अपराधों का हिसाब करें
> Translation: *AAP's ideology is like that of Pakistan. A conspiracy has been hatched against Punjab. Criminals should be held accountable through votes.* **(Hostile/ Demeaning)**
> ➢सुनने में आ रहा है की जिन आतंकियों को मृत्युदंड दिया गया है उनमे से एक आजमगढ़ का है और उसका बाप समाजवादी पार्टी में है। तो बात हुई ना सही की समाजवादी की लाल टोपी निर्दोष भारतीयों के रक्त से रंगा हुआ है।
> Translation: *It is being heard that one of the terrorists who has been given the death penalty is from Azamgarh and his father is associated with the Samajwadi Party. So it is correct to say that the red cap of Samajwadi Party is stained with blood of innocent Indians.* **(Misinformation)**

**No Hate Speech**: The characteristics of non-hate speech were defined as part of the annotation guidelines for a study on hate speech during the Indian election campaign. These include constructive criticism of political figures, policies, or parties, factual and informative content, lack of hostility, and absence of misinformation or fake news. Non-hate speech during the Indian election campaign also does not target specific groups of people based on their political beliefs or affiliations. These guidelines were provided to annotators to help them identify and label tweets as non-hate speech.

> ➢काम करते तो जनता खुद मन से वोट देते । अभी जनता जागरूक हो गइ है ।
> Translation: *If you had done your work, people would have voted with their heart. People have become aware now.* **(Constructive Criticism)**
> ➢कोविड से आपकी सुरक्षा के लिए मतदान केंद्रों पर होंगे कई इंतज़ाम। अब आप भी हो जाइये, मतदान के लिए तैयार।
> Translation: *Arrangements to prevent you from COVID has been done. Now, you should also be ready for voting.* **(No Misinformation)**

Labeling tweets for hate speech can be challenging (Thapa et al. 2022). Thus, a multi-phase approach was taken to ensure high-quality annotations. First, a pilot annotation of 100 tweets was conducted with all four team members to confirm their understanding of the instructions. The instructions were revised based on the feedback received to address any confusion. In the second phase, all four annotators were tasked with labeling 250 tweets to assess the clarity of the revised instructions. Finally, a group discussion was conducted to resolve any conflicts in the annotations. This multi-phase approach ensures consistency and accuracy in the annotation process, even when the content of the tweet is challenging. The inter-annotator agreement score, Fleiss's Kappa ($\kappa$), was 0.49 and 0.72 for pilot and final annotations, respectively.

### Dataset Statistics and Analysis

Our new dataset included 11,457 tweets, with 970 (8.46%) of them annotated as "Hate Speech" and 10,487 (91.53%) as "No Hate" (Table 1). Dataset statistics reflect a real-world scenario where most posts are neutral, and only a few contain hate speech.

| Labels | Tweets | Avg. Char | Avg. words |
| --- | --- | --- | --- |
| Hate | 970 | 158.37 (145.02) | 27.39 (25.37) |
| Non-Hate | 10487 | 150.06 (136.70) | 24.51 (22.69) |

Table 1: Dataset Statistics for "IEHate" dataset. The values in parentheses in average characters per tweet (Avg. Char) and average words per tweet (Avg. words) are calculated after preprocessing of text.

### Exploratory Data Analysis

The top 5 words in our entire dataset, as well as the categories of hate speech and non-hate speech, are shown in Table 2. In addition to the matching TF-IDF score for words, translations or transliterations are provided. Words with high TF-IDF scores are deemed to be more significant to the document. The terms Election (चुनाव), vote (मतदान), and Congress (कांग्रेस) have a high significance in majority of the courses, according to Tables 2. The histogram for the total number of words across both classes is displayed in Figure 1. The average number of words and characters per tweet for each class are shown in Table 2 as well. A relatively higher number of words in "Hate Speech" tweets show that people frequently engaged in lengthy discussions about how the current political system is flawed in "Hate" tweets.

## Experimental Results

We used various machine learning, deep learning, and state-of-the-art transformer-based methods to establish baselines. We further evaluated our test data using two human evaluators. They were made to label the tweets based on whether they contained hate or not. Instructions were provided to human evaluators. We employed accuracy, macro-mean-squared-error (MMAE), and F1-score (macro) as evaluation measures to determine each baseline's results.

|  | **All Posts** |  |  | **No Hate Speech Posts** |  |  | **Hate Speech Posts** |  |
|---|---|---|---|---|---|---|---|---|
| **Words** | **Translation** | **TF-IDF** | **Words** | **Translation** | **TF-IDF** | **Words** | **Translation** | **TF-IDF** |
| चुनाव | Election | 0.1327 | चुनाव | Election | 0.1389 | नहीं | No | 0.1290 |
| कांग्रेस | Congress | 0.0855 | मतदान | Vote | 0.0906 | कांग्रेस | Congress | 0.0959 |
| मतदान | Vote | 0.0843 | कांग्रेस | Congress | 0.0838 | भाजपा | BJP | 0.0869 |
| गोवा | Goa | 0.0757 | विधानसभा | Assembly | 0.0793 | पंजाब | Punjab | 0.0806 |
| विधानसभा | Assembly | 0.0741 | गोवा | Goa | 0.0782 | सरकार | Government | 0.0608 |

Table 2: Top-5 most frequent words in the overall dataset and also for each sub-classes viz. "Hate Speech" and "Non-Hate Speech" along with their TF-IDF values

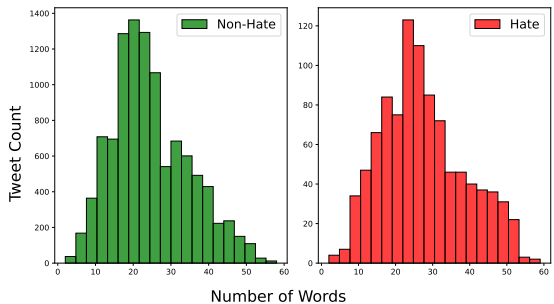

Figure 1: Histogram of the number of words used in Hate Speech and Non-Hate speech tweets

| Models | **Acc↑** | **MMAE↓** | **F1-Score↑** |
|---|---|---|---|
| Naive Bayes | 0.766 | 0.410 | 0.589 |
| Decision Tree | 0.849 | 0.438 | 0.568 |
| Logistic Regression | 0.820 | 0.410 | 0.604 |
| SVM | 0.916 | 0.303 | 0.533 |
| LSTM (Upscaled) | 0.892 | 0.350 | 0.639 |
| Bi-LSTM | 0.901 | 0.367 | 0.585 |
| Bi-LSTM + CNN | 0.873 | 0.391 | 0.616 |
| XLM-Roberta | 0.906 | 0.322 | 0.603 |
| BERT (HAM) | 0.913 | 0.272 | 0.706 |
| BERT (HCMAM) | 0.905 | 0.307 | 0.659 |
| RoBERTa | **0.923** | **0.230** | **0.725** |
| Human Evaluator-A | 0.959 | 0.126 | 0.847 |
| Human Evaluator-B | 0.968 | 0.087 | 0.881 |

Table 3: Baseline Results with different algorithms

**Machine Learning Algorithms**: In traditional machine learning algorithms, we used Naive Bayes (Rish et al. 2001), Decision Tree (Rokach and Maimon 2005), Logistic Regression (Stoltzfus 2011), Support Vector Machine (SVM) (Cortes and Vapnik 1995) with TF-IDF vectorizer.

**Deep Learning Algorithms**: In deep learning algorithms, we used Long-Short-Term-Memory (LSTM) (Hochreiter and Schmidhuber 1997), Bidirectional-LSTM, Bi-LSTM + Convolutional Neural Networks (CNN). We employed TensorFlow Tokenizer embedding for word embeddings.

**Transformer based algorithms**: In transformer-based models, we have used XLM-RoBERTa-base (Barbieri, Anke, and Camacho-Collados 2022) trained on 198M multilingual tweets, BERT (HAM: Hindi-abusive-MuRIL) (Das, Banerjee, and Mukherjee 2022) trained on Devanagari Hindi and was finetuned on MuRIL model, BERT (HCMAM: Hindi-codemixed-abusive-MuRIL) (Das, Banerjee, and Mukherjee 2022) and RoBERTa (hate-roberta-hasoc-hindi) (Velankar et al. 2021) model fine-tuned on Hindi HASOC Hate Speech Dataset 2021. All models were imported from huggingface[2] library.

## Results

Table 3 show the results for the classification of hate and non-hate speech. Among all the algorithms, the RoBERTa (multilingual) and BERT (HAM) models had a nearly equal best performance with an F1-score of 0.725 and 0.706, respectively. Among the machine learning algorithm, logistic regression performed the best with an F1-score of 0.604. We can see that the transformer-based models had better results than machine learning and deep learning models. The relatively lower F1-score of models compared to human evaluators emphasizes the need to develop more advanced and robust algorithms for hate speech detection in contexts like the Indian assembly election campaign.

## Conclusion and Future Work

This paper presents the IEHate dataset, which can be a valuable resource for developing and evaluating hate speech detection models in Indian election campaign discourse. The dataset consists of tweets in the Hindi language, which is a language that has not been studied extensively in hate speech detection. Despite the inherent subjectivity of annotations, the high inter-annotator agreement indicates a good level of consistency in labeling hate speech. Developing novel NLP models tailored to detecting hate speech in the context of Indian Election campaigns would be a promising area of exploration. Additionally, exploring hate speech toward detecting targets can be a promising direction. Overall, we believe that the IEHATE dataset will contribute to advancing the development of effective hate speech detection models, promoting a more inclusive and respectful online discourse in India.

---
[2] https://huggingface.co/